\documentclass[lettersize,journal]{IEEEtran}

\usepackage{amsmath,amsfonts,amssymb}
\usepackage{bm} 

\usepackage[ruled,linesnumbered]{algorithm2e}

\usepackage{array}
\usepackage{multirow}
\usepackage{booktabs}
\usepackage{threeparttable}
\usepackage{colortbl}
\usepackage{wasysym} 

\usepackage{graphicx}

\usepackage{url}
\usepackage{makecell}
\usepackage{orcidlink}
\usepackage{microtype}

\usepackage{cite}
\usepackage{hyperref}
\hypersetup{
 colorlinks = true,
 linkcolor = blue,
 filecolor = magenta,
 citecolor = blue,
 urlcolor = black
}
\usepackage[capitalise]{cleveref}

\begin{document}

\title{See What You Seek: Semantic Contextual Integration for Cloth-Changing Person Re-Identification}

\author{
 Xiyu~Han\orcidlink{0009-0006-9732-2939},
 Xian~Zhong\orcidlink{0000-0002-5242-0467},~\IEEEmembership{Senior~Member,~IEEE},
 Wenxin~Huang\orcidlink{0000-0001-8683-7327},~\IEEEmembership{Member,~IEEE},
 Xuemei~Jia\orcidlink{0003-1762-4365},
 Xiaohan~Yu\orcidlink{0000-0001-6186-0520},
 and~Alex~Chichung~Kot\orcidlink{0000-0001-6262-8125},~\IEEEmembership{Life~Fellow,~IEEE}

\thanks{Manuscript received May 17, 2025. This work was supported in part by the National Natural Science Foundation of China under Grants 62271361 and 62301213, the Hubei Provincial Key Research and Development Program under Grant 2024BAB039, and the Open Project Funding of the Hubei Key Laboratory of Big Data Intelligent Analysis and Application, Hubei University under Grant 2024BDIAA01.
(\textit{Corresponding author: zhongx@whut.edu.cn})}

\thanks{Xiyu Han and Xian Zhong are with the Sanya Science and Education Innovation Park, Wuhan University of Technology, Sanya 572025, and also with the Hubei Key Laboratory of Transportation Internet of Things, School of Computer Science and Artificial Intelligence, Wuhan University of Technology, Wuhan 430070, China (e-mail: hanxy@whut.edu.cn; zhongx@whut.edu.cn).}%

\thanks{Wenxin Huang is with the Hubei Key Laboratory of Big Data Intelligent Analysis and Application, School of Computer Science and Information Engineering, Hubei University, Wuhan 430062, China (e-mail: wenxinhuang\_wh@163.com).}%

\thanks{Xuemei Jia is with the National Engineering Research Center for Multimedia Software, School of Computer Science, Wuhan University, Wuhan 430072, China (e-mail: jiaxuemeiL@whu.edu.cn).}%


\thanks{Xiaohan Yu is with the School of Computing, Macquarie University, Sydney, NSW 2109, Australia (e-mail: xiaohan.yu@mq.edu.au).}%

\thanks{Alex Chichung Kot is with the Rapid-Rich Object Search Lab, School of Electrical and Electronic Engineering, Nanyang Technological University, Singapore 639798 (e-mail: eackot@ntu.edu.sg).}

}

\markboth{IEEE Transactions on Circuits and Systems for Video Technology, 2025}%
{Shell \MakeLowercase{\textit{et al.}}: A Sample Article Using IEEEtran.cls for IEEE Journals}


\maketitle

\begin{abstract}
Cloth-changing person re-identification (CC-ReID) aims to match individuals across surveillance cameras despite variations in clothing. Existing methods typically mitigate the impact of clothing changes or enhance identity (ID)-relevant features, but they often struggle to capture complex semantic information. In this paper, we propose a novel prompt learning framework \emph{Semantic Contextual Integration} (SCI), which leverages the visual-textual representation capabilities of CLIP to reduce clothing-induced discrepancies and strengthen ID cues. Specifically, we introduce the \emph{Semantic Separation Enhancement} (SSE) module, which employs dual learnable text tokens to disentangle clothing-related semantics from confounding factors, thereby isolating ID-relevant features. Furthermore, we develop a \emph{Semantic-Guided Interaction Module} (SIM) that uses orthogonalized text features to guide visual representations, sharpening the focus of the model on distinctive ID characteristics. This semantic integration improves the discriminative power of the model and enriches the visual context with high-dimensional insights. Extensive experiments on three CC-ReID datasets demonstrate that our method outperforms state-of-the-art techniques. The code will be released at \url{https://github.com/hxy-499/CCREID-SCI}.

\end{abstract}


\section{Introduction}

\IEEEPARstart{P}{erson} re-identification (ReID) is the task of matching individuals across non-overlapping cameras over time, with important applications in video surveillance and smart city systems~\cite{action,search,alexkot,AFL}. Traditional ReID methods~\cite{yxhsearch,hxyyxh} exploit the appearance of clothing, \textit{e.g.}, texture and color, to distinguish identities (IDs), but their performance degrades when subjects change clothing. This limitation motivates the development of cloth-changing ReID (CC-ReID) techniques that remain robust under clothing variations.

\begin{figure}[t]
	\centering
	\includegraphics[width = \linewidth]{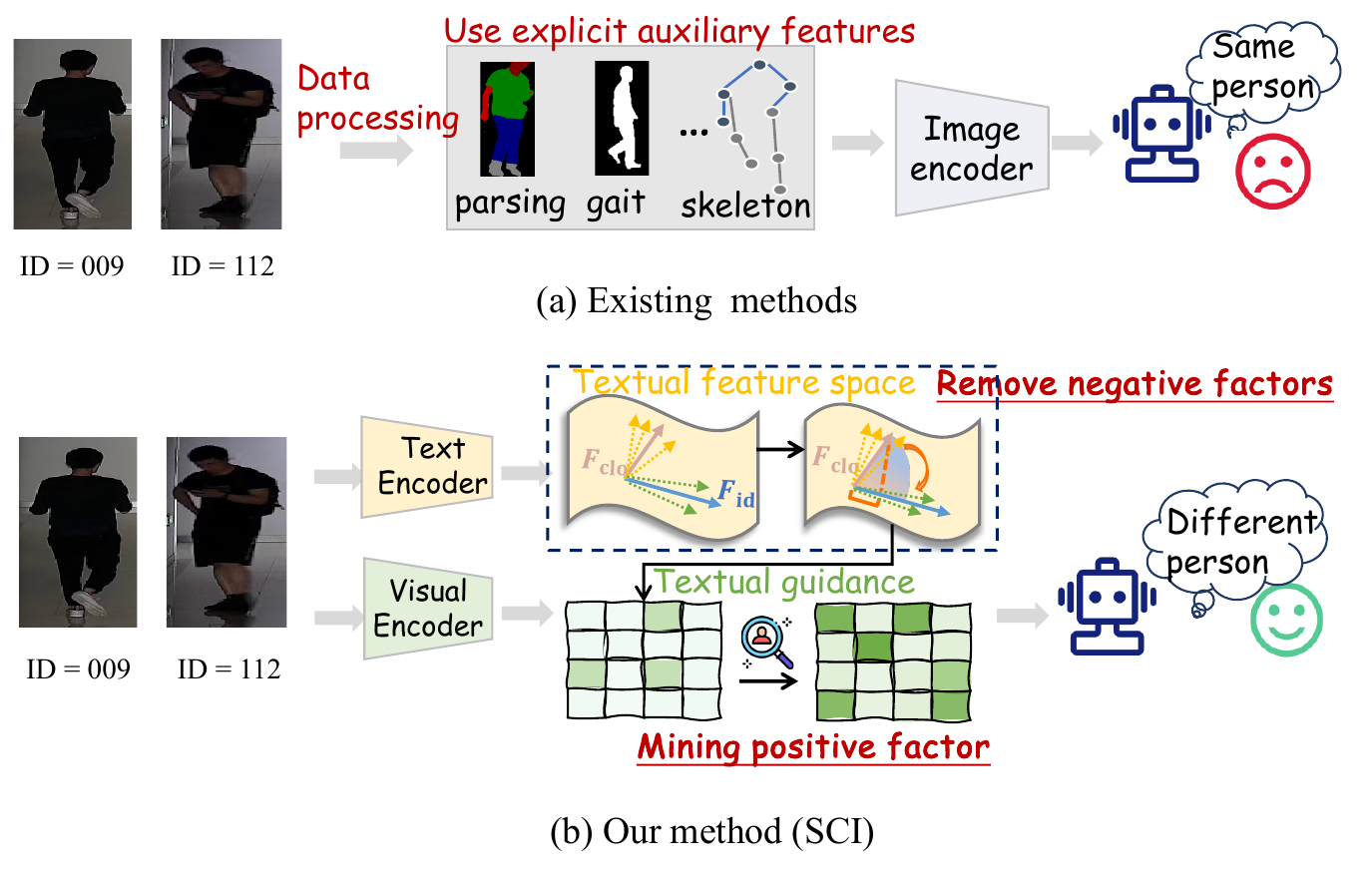}
	\caption{\textbf{Comparison of traditional methods and our SCI approach.} (a) Traditional methods rely on parsing, gait analysis, skeleton extraction, and data augmentation to suppress clothing effects, incurring significant preprocessing overhead. (b) Our SCI approach directly removes clothing bias within the model and exploits inherent ID-related features from images.}
	\label{fig_motivation}
\end{figure}

Existing CC-ReID approaches can be grouped into two categories: those that suppress clothing cues and those that enhance ID-relevant features (see \cref{fig_motivation}(a)). The first category seeks to minimize the influence of clothing: CAL~\cite{cal} employs adversarial learning to penalize clothing-based predictions in the RGB modality, and AIM~\cite{aim} uses causal intervention through a dual-branch model to mitigate clothing bias. The second category injects auxiliary ID cues, such as pose, gait, or human parsing, to strengthen discriminative features. For example, GI-ReID~\cite{gireid} leverages gait information to learn clothing-agnostic representations, while SCNet~\cite{scnet} imposes semantic consistency constraints using a human parsing network.

Although these methods partially address clothing changes, they typically focus on minimizing negative clothing effects or relying on explicit cues such as body contours. We argue that visual representations inherently contain both negative factors (\textit{e.g.}, clothing styles that frequently change and can confuse the model) and positive factors (\textit{e.g.}, stable attributes such as hair, glasses, or backpacks that aid ID discrimination). As illustrated in \cref{fig_motivation}(b), our goal is to remove negative influences while making implicit positive factors more explicit.

Recent advances in vision-language learning, notably Contrastive Language-Image Pre-training (CLIP)~\cite{clip}, have demonstrated strong abilities to bridge visual data and natural language, yielding context-aware representations beneficial for many downstream tasks~\cite{clipdection,clippasso}. In ReID, methods like CLIP-ReID~\cite{clipreid} and Instruct-ReID~\cite{Instruct-ReID} show that aligning images with descriptive language captures rich semantics beyond appearance alone. Inspired by these successes, we leverage prompt learning to extract high-dimensional semantic cues from CLIP, thereby amplifying positive factors without additional data preprocessing.

In this paper, we introduce the \emph{Semantic Contextual Integration} (SCI) framework for CC-ReID. SCI uses CLIP to capture rich semantic features and includes a \emph{Semantic Separation Enhancement} (SSE) module isolates clothing-related negative factors at the text level while preserving key positive semantics for ReID. Specifically, SSE employs dual learnable text tokens to disentangle confounding semantics (both positive and negative) from clothing semantics, then orthogonalizes the token embeddings to filter out clothing factors. The refined text features guide visual encoding via our \emph{Semantic-Guided Interaction Module} (SIM), focusing the model’s attention on ID-relevant cues beyond explicit contours. By integrating CLIP’s multimodal strengths, SCI enriches visual representations with high-dimensional semantic context and achieves superior performance on three standard CC-ReID benchmarks.

Our contributions are summarized threefold:

\begin{itemize}
	\item We propose the \emph{Semantic Contextual Integration} (SCI) framework to leverage semantic information in CC-ReID, removing negative factors, emphasizing implicit positive elements, and refining visual representations.

	\item We introduce the \emph{Semantic Separation Enhancement} (SSE) module to filter and refine text-level features, improving the model’s ability to isolate key semantic information.

	\item We design the \emph{Semantic-Guided Interaction Module} (SIM) to guide visual representations using refined text features, enhancing multi-modal integration and alignment.


\end{itemize}


\begin{figure*}[t]
	\centering
	\includegraphics[width = 0.95\linewidth]{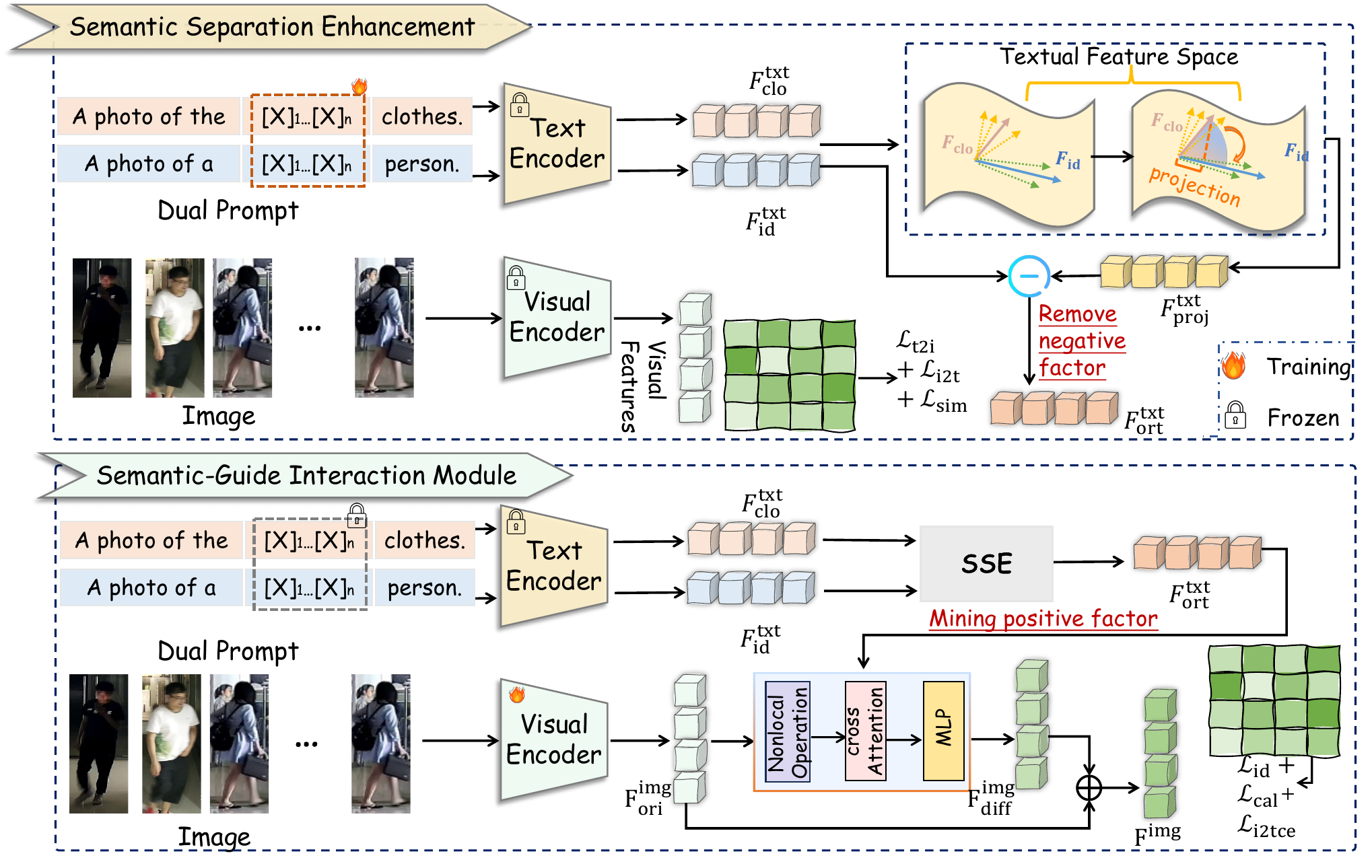}
	\caption{\textbf{Framework of the proposed SCI, comprising two key components:} the Semantic Separation Enhancement (SSE) module and the Semantic-Guided Interaction Module (SIM). SSE mitigates clothing bias by removing negative semantic factors, while SIM employs the refined text features to guide visual representations, strengthening cross-modal interaction.}
	\label{fig_framework}
\end{figure*}

\section{Related Work}

\subsection{Cloth-Consistent Person ReID}

Person ReID under consistent clothing conditions has been extensively studied~\cite{cmt_tcsvt}. These methods exploit clothing appearance, such as color and texture, to extract discriminative features. Feature representation learning in this context can be divided into three categories: global features, local features, and auxiliary information-based features. Global methods extract a single feature vector per image~\cite{global}, while local methods aggregate part-based representations to address misalignment (\textit{e.g.}, human parsing~\cite{parse} or horizontal partitioning~\cite{pcb}). Auxiliary-information approaches incorporate extra cues, such as semantic attributes~\cite{tip_attribute} or synthesized samples~\cite{augmentation}, to enrich context. Strong baselines in this domain include AGW~\cite{agw} and TransReID~\cite{transreid}. ISP~\cite{isp} further refines alignment by locating body parts and carried items at the pixel level. 
Moreover, since person ReID is inherently formulated as a ranking problem, class imbalance can also have a significant impact on performance. Therefore, some works propose loss function modifications to  handle data imbalance during training. For example, DRSL~\cite{yuanxin_loss1} optimizes the ReID model by enforcing positives to rank ahead of negatives based on distance, and further sorting positives by similarity score. Parameterized RV Loss~\cite{yuanxin_loss2} jointly optimizes retrieval and verification tasks by aligning loss functions with evaluation metrics, enabling automatic loss function search. 

Although these techniques handle variations in pose, background, and viewpoint, they assume identical clothing across views. Consequently, their performance degrades when subjects change clothing, motivating research on CC-ReID.

\subsection{Cloth-Changing Person ReID}

Several CC-ReID datasets have been proposed, \textsc{PRCC}~\cite{prcc}, \textsc{LTCC}~\cite{ltcc}, \textsc{COCAS+}~\cite{cocas_tcsvt}, and \textsc{VC-Clothes}~\cite{vc}, to evaluate models under clothing variations. The primary challenge is to learn features that remain reliable despite clothing changes. Existing CC-ReID methods follow three main strategies:

\subsubsection{Auxiliary Soft-Biometric Cues}
CAMC~\cite{camc} integrates body-shape semantics into ID features, and FSAM~\cite{fsam} extracts fine-grained shape information to complement clothing-independent cues. M2Net~\cite{m2net} uses contour and parsing maps for appearance-robust features, PGAL~\cite{pose_pgal} aligns keypoints via pose estimation, FLAG~\cite{flag} is proposed to explicitly extract appearance and gait information, and can be integrated with most existing video-based ReID methods, and CVSL~\cite{cvsl} jointly learns body-shape embeddings and appearance features.

\subsubsection{Feature Disentanglement}
CAL~\cite{cal} employs adversarial loss to suppress clothing-related features, and AIM~\cite{aim} uses causal intervention to remove clothing bias. DCR-ReID~\cite{DCR-ReID_tcsvt} disentangles and reconstructs feature components, LDF~\cite{ldf} separates ID, clothing, and unrelated factors via GANs, 3DInvarReID~\cite{3DInvarReID} disentangles features and reconstructs two-layer 3D body shapes, FIRe$^2$~\cite{Fire2} augments images with fine-grained attributes, and MAL-F~\cite{MAL-F} learns invariant features from RGB, grayscale, and contour inputs using a ResTNet backbone.

\subsubsection{Data Augmentation and Adaptation}
Pos-Neg~\cite{posneg} and CCFA~\cite{ccfa} augment clothing color and texture diversity; RCSANet~\cite{rcsanet} constructs explicit clothing-status embeddings to bolster feature robustness; MCSC~\cite{meta} applies meta-learning to address clothing-distribution shifts; Zhao \textit{et al.}~\cite{graph} model clothing changes as fine-grained domain shifts via graph relations; and DCLR~\cite{dclr} synthesizes multi-clothing images via diffusion and merges them into training data.

Despite these advances, most CC-ReID methods rely solely on visual inputs, limiting their semantic depth. In this work, we integrate visual and textual modalities via CLIP to enrich semantic understanding and improve cloth-changing ReID.

\subsection{Vision-Language Learning}

Vision-language pre-training (VLP) trains models to align images with text, improving downstream visual tasks. CLIP~\cite{clip} uses paired image and text encoders to learn a shared embedding space via contrastive learning, benefiting applications such as captioning and classification.

Prompt learning extends CLIP by making context tokens learnable. CoOp~\cite{coop} transforms prompt words into trainable vectors, CoCoOp~\cite{conditionprompt} generates input-conditional tokens per image, and DenseCLIP~\cite{denseclip} applies pixel-text matching for dense prediction.

In person ReID, CLIP-ReID~\cite{clipreid} introduces ID-specific tokens and a two-stage training scheme. CCLNet~\cite{cclnet} adapts prompt learning for unsupervised visible-infrared ReID. CSDN~\cite{csdn} employs bimodal descriptions to align visible and infrared features. RGANet~\cite{RGANet_occluded} uses CLIP to locate informative body parts for occluded ReID. VGSG~\cite{VGSG_group} groups text features semantically to address fine-grained misalignment. 
CCAFL~\cite{clip_camera} integrates CLIP-generated textual information into a novel semi-supervised framework to actively learn camera-agnostic pedestrian features. MIPL~\cite{tip_mutinformation} guides CC-ReID with multiple common prompts, and CCAF~\cite{arxiv_555} leverages CLIP’s knowledge to learn fine-grained, clothing-independent semantics.

Although CLIP is not specifically pre-trained for ReID, its large-scale image-text training captures high-level semantics (\textit{e.g.}, body shape and context) remain invariant to clothing. We build on this by proposing CLIP-CCReID, which uses prompt learning to mitigate clothing changes without auxiliary models.

\section{Proposed Method}

\subsection{Preliminaries and Overview}

The large-scale vision-language model CLIP~\cite{clip} learns joint image-text representations via a contrastive objective. It comprises a visual encoder $\mathcal{I}(\cdot)$ and a text encoder $\mathcal{T}(\cdot)$, whose embedding spaces are aligned during pre-training. 

For downstream classification, CoOp~\cite{coop} refines CLIP’s zero-shot prompts by introducing learnable context tokens. Specifically, given a template such as 
\texttt{[$V$]\textsubscript{1} [$V$]\textsubscript{2} $\dots$ [$V$]\textsubscript{M} [CLS]}
where each [$V$]\textsubscript{i} ($i=1,\dots,M$) is a trainable vector and [CLS] is a class token, CLIP computes image-text similarities in the shared embedding space and predicts the class with maximal similarity. 

Building on these ideas, we leverage CLIP and CoOp to extract rich semantic prompts for CC-ReID (see \cref{fig_framework}), enabling robust ID modeling under clothing changes.

\begin{algorithm}[t]
 \caption{Semantic Separation Enhancement Module}
 \DontPrintSemicolon 
 \SetAlgoVlined 
 \SetKwInOut{KwIn}{Input}
 \SetKwInOut{KwOut}{Output}
 \label{alg_s2e}
 \small
 \KwIn{Dual textual learnable vectors $\mathrm{prompt_\mathrm{id}}$, $\mathrm{prompt_\mathrm{clo}}$, and person images.}
 \KwOut{Optimized text representations ${F^\mathrm{txt}_\mathrm{ort}}$}
 
 \For{each epoch $e$ from $1$ to epochs}{
 \For{each batch in trainloader}{
 Extract image features $F_\mathrm{img}$ using the frozen visual encoder $\mathcal{I}(\cdot)$ 
 Train context vectors for dual prompts $\mathrm{prompt_\mathrm{id}}$ and $\mathrm{prompt_\mathrm{clo}}$ with \cref{eq_learncontext1} and \cref{eq_learncontext2} 
 Compute text features ${F^\mathrm{txt}_\mathrm{id}}$ and ${F^\mathrm{txt}_\mathrm{clo}}$ for ID and clothing using \cref{eq_computefea} 
 Project ${F^\mathrm{txt}_\mathrm{clo}}$ onto ${F^\mathrm{txt}_\mathrm{id}}$ using \cref{eq_proj} 
 Perform orthogonalization to obtain ${F^\mathrm{txt}_\mathrm{ort}}$ using \cref{eq_ort} 
 Compute $\mathcal{L}_\mathrm{sim}$ loss to regulate semantic separation using \cref{eq_lambda} 
}
}
 \Return{${F^\mathrm{txt}_\mathrm{ort}}$}
\end{algorithm}

\subsection{Semantic Separation Enhancement}

CLIP cannot explicitly ignore unwanted semantics (\textit{e.g.}, clothing) via negative prompts. To address this, we propose the Semantic Separation Enhancement (SSE) module, which isolates and removes clothing-related components from the high-dimensional text features, preserving only ID-relevant semantics. The overall procedure is detailed in \cref{alg_s2e}.

We define two learnable prompts:
\begin{equation}
	\mathrm{prompt_\mathrm{id}} = \mathrm{A~photo~of~a~[X]_1~[X]_2~\dots [X]_M~person.}
\label{eq_learncontext1}
\end{equation}
\begin{equation}
	\mathrm{prompt_\mathrm{clo}} = \mathrm{A~photo~of~the~[X]_1~[X]_2~\dots [X]_M~clothes.}
\label{eq_learncontext2}
\end{equation}
where $[X]_i$ ($i = 1, \dots, M$) are learnable tokens initialized randomly, and $M$ denotes the number of tokens. In the first training stage, we use the pre-trained visual encoder $\mathcal{I}(\cdot)$ and text encoder $\mathcal{T}(\cdot)$ to extract image and dual text features, freezing the encoders' parameters to focus on optimizing the text tokens $[X]_i$. This allows us to learn contextual representations by updating the text tokens, thereby acquiring distinct textual representations for each ID and their clothing:
\begin{equation}
	F^\mathrm{txt}_\mathrm{id} = \mathcal{T} \left(\mathrm{prompt_\mathrm{id}} \right), \quad
	F^\mathrm{txt}_\mathrm{clo} = \mathcal{T} \left(\mathrm{prompt_\mathrm{clo}} \right),
\label{eq_computefea}
\end{equation}
where $F^\mathrm{txt}_\mathrm{id}$ and $F^\mathrm{txt}_\mathrm{clo}$ represent the ID-specific text features and clothing text features, respectively.

After extracting the dual text representations, we project $F^\mathrm{txt}_\mathrm{clo}$ onto $F^\mathrm{txt}_\mathrm{id}$ to reduce the influence of clothing and emphasize the implicit positive factors. This step clarifies the relation between clothing representations and ID representations, thereby minimizing the impact of clothing variations. The computation process is as follows:
\begin{equation}
	F^\mathrm{txt}_\mathrm{proj}= \frac{F^\mathrm{txt}_\mathrm{clo} F^\mathrm{txt}_\mathrm{id}}{\left\|F^\mathrm{txt}_\mathrm{id} \right\|^2} F^\mathrm{txt}_\mathrm{id},
\label{eq_proj}
\end{equation}
We then perform orthogonalization by subtracting the projection $F^\mathrm{txt}_\mathrm{proj}$ of the clothing feature $F^{\mathrm{txt}}_{\mathrm{clo}}$ onto the $F^{\mathrm{txt}}_{\mathrm{id}}$. This operation removes the clothing component that is aligned with the ID direction. This process ensures that the features $F^\mathrm{txt}_\mathrm{ort}$ are aligned with the positive features while removing the influence of negative factors. The orthogonalization process is:
\begin{equation}
	F^\mathrm{txt}_\mathrm{ort} = F^\mathrm{txt}_\mathrm{id} - F^\mathrm{txt}_\mathrm{proj},
\label{eq_ort}
\end{equation}
where $F^\mathrm{txt}_\mathrm{ort}$ is after removing the part that is related to clothing, ensuring the feature focuses on positive information while reducing interference from clothing.

To optimize the text features, we design a loss function that maximizes the similarity between $F^\mathrm{txt}_\mathrm{ort}$ and $F^\mathrm{txt}_\mathrm{id}$, while minimizing the similarity with $F^\mathrm{txt}_\mathrm{clo}$. This enables the model to better capture positive factors without being affected by negative aspects, such as clothing changes. The similarity calculations are:
\begin{equation}
	\mathrm{sim_\mathrm{id}}(i) = \mathrm{mean} \left(\cos \left(F^\mathrm{txt}_\mathrm{ort}(i), F^\mathrm{txt}_\mathrm{id}(i) \right) \right),
\end{equation}
\begin{equation}
	\mathrm{sim_\mathrm{clo}}(i) = \mathrm{mean} \left(\cos \left(F^\mathrm{txt}_\mathrm{ort}(i), F^\mathrm{txt}_\mathrm{clo}(i) \right) \right),
\end{equation}
where $\mathrm{mean}(\cdot)$ denotes the average value, and $\cos(\cdot)$ represents the cosine similarity function. 
The loss function is as follows:
\begin{equation}
	\mathcal{L}_\mathrm{sim}(i) = \lambda_1 \left(1 - \mathrm{sim_\mathrm{id}}(i) \right) + \lambda_2~\mathrm{sim_\mathrm{clo}}(i),
 \label{eq_lambda}
\end{equation}
where $\lambda_1$ and $\lambda_2$ are weighting factors controlling the contributions of ID and clothing similarities.

Finally, we calculate the image-to-text contrastive loss $\mathcal{L}_\mathrm{i2t}$:
\begin{equation}
	\mathcal{L}_\mathrm{i2t}(i) = -\log \frac{\exp s \left(V_i, F^\mathrm{txt}_\mathrm{ort}(i) \right)}{\sum_{k = 1}^N \exp s \left(V_i, F^\mathrm{txt}_\mathrm{ort}(k) \right)},
\end{equation}
where $V_i$ and $F^\mathrm{txt}_\mathrm{ort}(i)$ are paired visual and text embeddings, $s(\cdot, \cdot)$ denotes the similarity function, and $N$ is the batch size. Since multiple images in a batch may belong to the same ID, meaning there may be multiple positive samples, the text-to-image contrastive loss $\mathcal{L}_\mathrm{t2i}(y_i)$ is calculated as:
\begin{equation}
	\mathcal{L}_\mathrm{t2i}(y_i) = \frac{-1}{|P(y_i)|} \sum_{p \in P(y_i)} \log \frac{\exp s \left(V_p, T_{y_i} \right)}{\sum_{k = 1}^N \exp s \left(V_k, T_{y_i} \right)},
\end{equation}
where $P(y_i)$ is the set of all positive indices for $F^\mathrm{txt}_\mathrm{ort}(y_i)$ in the batch, and $T_{y_i}$ is the text embedding corresponding to label $y_i$. 
Therefore, the overall loss function is:
\begin{equation}
	\mathcal{L}_\mathrm{prompt} = \sum_{i = 1} \left(\mathcal{L}_\mathrm{i2t} + \mathcal{L}_\mathrm{t2i} + \mathcal{L}_\mathrm{sim} \right).
\end{equation}




\begin{algorithm}[t]
	\caption{Semantic-Guided Interaction Module}
	\DontPrintSemicolon 
	\SetAlgoVlined 
	\SetKwInOut{KwIn}{Input}
	\SetKwInOut{KwOut}{Output}
	\label{alg_sim}
	\small
	\KwIn{Dual textual learnable vectors $\mathrm{prompt_\mathrm{id}}$, $\mathrm{prompt_\mathrm{clo}}$, and person images.}
	\KwOut{Optimized text representations ${F^\mathrm{txt}_\mathrm{ort}}$}
 
	\For{each epoch $e$ from $1$ to epochs}{
	\For{each batch in trainloader}{
	Extract dual text features $F_\mathrm{txt}$ using the frozen text encoder $\mathcal{T}(\cdot)$ 
	Compute orthogonalized text features ${F^\mathrm{txt}_\mathrm{ort}}$ 
	Extract image features $F_\mathrm{img}$ using the training visual encoder $\mathcal{I}(\cdot)$ 
	Enhance image features with contextual information using \cref{eq_nonlocal} 
	Compute semantic-guided image features with \cref{eq_transdec1} and \cref{eq_transdec2} 
	Refine image features based on ${F^\mathrm{txt}_\mathrm{ort}}$ using \cref{eq_finalimg} 
	Apply loss functions to regulate the visual encoder training using \cref{eq_stage2loss} 
}
}
\end{algorithm}

\subsection{Semantic-Guide Interaction Module}

Previous work~\cite{cal,clipreid} shows that reducing clothing bias in visual or textual features improves performance in cloth-changing scenarios. However, these methods typically treat visual and textual branches independently, lacking interaction. Our method refines visual representations by leveraging clothing-irrelevant textual features, enhancing the interaction between visual and textual branches. This integration introduces more robust and invariant descriptors into the visual processing, improving the model's ability to identify individuals across clothing changes. The algorithm is summarized in \cref{alg_sim}.

\begin{figure}[t]
	\centering
	\includegraphics[width = \linewidth]{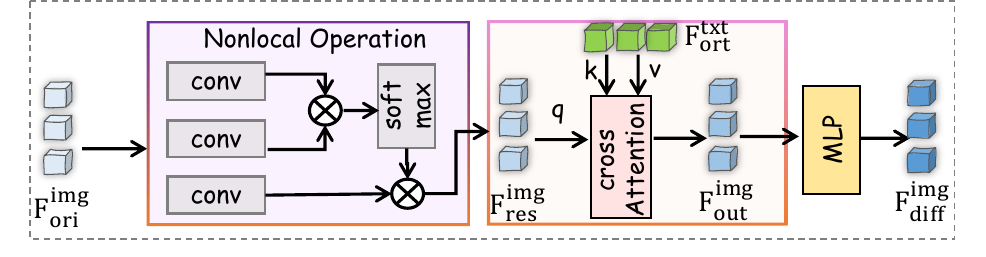}
	\caption{\textbf{Illustration of the SIM process.} Textual information refines visual feature extraction to align features with the relevant semantic context.}
	\label{fig_transcoder}
\end{figure}

As illustrated in \cref{fig_transcoder}, we employ a Transformer decoder~\cite{attentionisall} to model interactions between visual and textual data. To better capture long-range dependencies within the visual domain, we replace the conventional self-attention mechanism with a non-local operation~\cite{nonlocal}, which allows the model to aggregate information from spatially distant but semantically relevant regions. This capability is particularly important for handling variations in clothing. 
During this stage, we update the visual encoder while keeping other components frozen. The computation is as follows:
\begin{equation}
	F^{\mathrm{img}}_{\mathrm{con}} = \frac{\theta \left(F^{\mathrm{img}}_{\mathrm{ori}} \right) \phi \left(F^{\mathrm{img}}_{\mathrm{ori}} \right)}{N} g \left(F^{\mathrm{img}}_{\mathrm{ori}} \right),
\end{equation}
where $F^{\mathrm{img}}_{\mathrm{ori}}$ is the input feature map, $\theta(\cdot)$, $\phi(\cdot)$, and $g(\cdot)$ are linear transformations, and $N$ is the number of elements in the feature map. $F^{\mathrm{img}}_{\mathrm{con}}$ captures global context information.

The final visual feature map is:
\begin{equation}
	F^{\mathrm{img}}_{\mathrm{res}} = W \left(F^{\mathrm{img}}_{\mathrm{con}} \right) + F^{\mathrm{img}}_{\mathrm{ori}},
 \label{eq_nonlocal}
\end{equation}
where $W$ contains learnable parameters. $F^{\mathrm{img}}_{\mathrm{res}}$ is the final output feature map obtained by adding this context-aware feature to the original feature map.

To refine visual representations with semantic guidance, we leverage textual embeddings to generate attention weights over visual features. This allows the model to selectively emphasize semantically relevant regions in the visual space, leading to more discriminative and context-aware visual representations:
\begin{equation}
	F^{\mathrm{img}}_{\mathrm{out}} = F^{\mathrm{img}}_{\mathrm{res}} + \mathrm{softmax} \left(\frac{F^{\mathrm{img}}_\mathrm{res} {F^{\mathrm{txt}}_{\mathrm{ort}}}^T}{\sqrt{d_k}} \right) F^{\mathrm{txt}}_{\mathrm{ort}},
 \label{eq_transdec1}
\end{equation}
\begin{equation}
	F^{\mathrm{img}}_{\mathrm{diff}} = \mathrm{MLP} \left( \mathrm{LayerNorm} \left( F^{\mathrm{img}}_{\mathrm{out}} \right) \right), 
 \label{eq_transdec2}
\end{equation}
where $d_k$ is the dimensionality of the keys, $\mathrm{LayerNorm}$ is used to normalize the input features to stabilize training, and $\mathrm{MLP}$ denotes a two-layer feed-forward network. This integrates visual features under the guidance of textual embeddings, and enables the model to extract the most informative visual cues aligned with the semantic content of the text.

Finally, we update the visual features. This process ensures that the visual features are enriched with relevant semantic information from the text, leading to more robust and contextually aware visual representations:
\begin{equation}
	F^{\mathrm{img}} = F^{\mathrm{img}}_{\mathrm{ori}} + \alpha F^{\mathrm{img}}_{\mathrm{diff}},
 \label{eq_finalimg}
\end{equation}
where $\alpha \in \mathbb{R}^C$ is a learnable parameter controlling the degree of refinement applied to the visual features.

This module adopts an interaction-to-alignment approach, where text embeddings guide the visual encoding process. By emphasizing positive cues and mitigating negative ones, this method improves the model’s ability to consistently recognize individuals across different clothing scenarios.

For the objective function, we incorporate cross-entropy loss $\mathcal{L}_\mathrm{id}$ and clothes-based adversarial loss $\mathcal{L}_\mathrm{cal}$~\cite{cal} to optimize the visual encoder:
\begin{equation}
	\mathcal{L}_\mathrm{id} = -\sum_{i = 1}^N y^i \log \left(p_\mathrm{id} \left(y^i \mid x^i \right) \right),
\end{equation}
where $y^i$ is the true label for the $i$-th sample, and $p_\mathrm{id} \left( y^i \mid x^i \right)$ is the predicted probability of the true label $y^i$. The clothes-based adversarial loss is expressed as:
\begin{equation}
	\small
	\mathcal{L}_\mathrm{cal} = -\sum_{i = 1}^N \sum_{c = 1}^{N_C} q (c) \log \frac{\exp \left(f_i \varphi_c / \tau \right)}{\exp \left(f_i \varphi_c / \tau \right) + \sum_{j \in S_i^-} \exp \left(f_i \varphi_j / \tau \right)},
\end{equation}
where $N_C$ is the number of clothing categories, $\varphi_c$ is the clothes classifier, $q(c)$ is the weight for the $c$-th class, $f_i$ is the feature of sample $i$, $\tau$ is a temperature parameter, and $S_i^-$ is the set of clothes classes with the same
ID. The objective of $\mathcal{L}_\mathrm{cal}$ is to encourage the backbone to extract clothes-irrelevant features by penalizing its predictive power regarding clothing. 

To fully leverage the capabilities of CLIP, we calculate the image-to-text cross-entropy loss $\mathcal{L}_\mathrm{i2tce}$ as:
\begin{equation}
	\mathcal{L}_\mathrm{i2tce}(i) = \sum_{k = 1}^N -q_k \log \frac{\exp s(V_i, F^\mathrm{txt}_\mathrm{ort}(i))}{\sum_{k = 1}^N \exp s(V_i, F^\mathrm{txt}_\mathrm{ort}(k))},
\end{equation}
where label smoothing is applied to $q_k$. The total loss for the SIM module is:
\begin{equation}
	\mathcal{L} = \mathcal{L}_\mathrm{id} + \mathcal{L}_\mathrm{cal} + \mathcal{L}_\mathrm{i2tce}.
\label{eq_stage2loss}
\end{equation}
By jointly optimizing $\mathcal{L}_\mathrm{id}$, $\mathcal{L}_\mathrm{cal}$, and $\mathcal{L}_\mathrm{i2tce}$, the model captures ID features from easy samples (same clothes) and progressively learns to distinguish hard cases (same ID, different clothes) while aligning visual and textual representations. The weights of loss terms are basic terms~\cite{cal,clipreid} and set as 1.

\section{Experimental Results}

\subsection{Datasets and Evaluation Metrics}

We evaluate our method on three standard cloth-changing datasets: \textsc{LTCC}~\cite{ltcc}, \textsc{PRCC}~\cite{prcc}, and \textsc{VC-Clothes}~\cite{vc}, following the protocols of CAL~\cite{cal} and AIM~\cite{aim}. Additionally, to validate generalization, we test on two conventional ReID datasets: \textsc{Market1501}~\cite{market1501} and \textsc{MSMT17}~\cite{msmt17}. 

\textsc{LTCC} contains 17,138 images of 152 IDs captured by 12 cameras. The training set comprises 9,576 images of 77 IDs; the testing set includes 7,543 images (493 queries and 7,050 gallery) of 75 IDs. This long-term dataset features frequent clothing changes and varied environmental conditions, with each ID wearing between two and fourteen outfits. 

\textsc{PRCC} consists of 33,698 images of 221 IDs from three cameras. Its training set has 22,898 images of 150 IDs, and the testing set has 10,800 images of 71 IDs. Each person appears in two outfits: cameras A and B share the same clothing, while camera C uses a different outfit. 

\textsc{VC-Clothes} is a synthetic dataset generated via the GTA5 engine. It contains 9,449 images of 256 IDs across four camera views, with 1,241 distinct clothing items distributed between training and testing sets. 

\textsc{Market1501} includes 12,936 training images of 751 IDs and 19,732 testing images of 750 IDs, captured by six cameras under the single-query evaluation protocol. 

\textsc{MSMT17} comprises 32,621 training images of 1,041 IDs and 51,027 testing images of 11,659 IDs across 15 cameras, it is one of the largest and most challenging ReID benchmarks. 

We employ the cumulative matching characteristic (CMC) curve and mean average precision (mAP) as evaluation metrics. Rank-$k$ in CMC measures the probability of the correct match appearing among the top-$k$ retrieved images, while mAP reflects the overall retrieval performance across all queries. 

Experiments are conducted under three protocols: 
\textit{1) General setting:} exclude same-ID and same-camera samples, using both cloth-changing and clothing-consistent pairs;
\textit{2) Same-clothes setting:} exclude same-ID and same-camera samples, using only clothing-consistent pairs;
\textit{3) Cloth-changing setting:} exclude same-ID, same-camera and same-clothing samples, using only cloth-changing pairs.

\subsection{Implementation Details}

We adopt two backbones for feature extraction: a modified ResNet-50 and ViT-B/16 pre-trained on CLIP. All images are resized to $384 \times 192$ pixels, with a batch size of 64. We use the Adam optimizer~\cite{adam} and apply random horizontal flipping, cropping, and erasing~\cite{randomearse} for data augmentation. 

For the CNN branch, a global attention pooling layer to reduce the feature dimension from 2048 to 1024, matching the text feature dimension (scaled from 512 to 1024). For the transformer branch, we employ 12 transformer layers with hidden size 768; a linear projection reduces the image feature dimension from 768 to 512 to align with the text features. 

Training proceeds in two stages: 
\textit{1) Stage 1 (prompt learning):} train two prompt learners for 60 epochs (ResNet-50) or 120 epochs (ViT-B/16) with an initial learning rate of $3.5 \times 10^{-4}$ and cosine decay. 
\textit{2) Stage 2 (visual encoder fine-tuning):} train the visual encoder $\mathcal{I}(\cdot)$ for 120 epochs (ResNet-50) or 30 epochs (ViT-B/16), starting at $3.5 \times 10^{-4}$ and reducing by a factor of 10 at epochs 40 and 70. 

All experiments are implemented in PyTorch and run on a single NVIDIA A100 GPU. 

\subsection{Comparison with State-of-the-Art Methods}

We compare SCI against state-of-the-art methods on \textsc{LTCC}, \textsc{PRCC} and \textsc{VC-Clothes}, including general ReID approaches (ISP~\cite{isp}, CLIP-ReID~\cite{clipreid}, and Instruct-ReID~\cite{Instruct-ReID}) and specialized cloth-changing methods (FSAM~\cite{fsam}, RCSANet~\cite{rcsanet}, CAL~\cite{cal}, GI-ReID~\cite{gireid}, M2Net~\cite{m2net}, AIM~\cite{aim}, Pos-Neg~\cite{posneg}, ACID~\cite{tip_win}, 3DInvarReID~\cite{3DInvarReID}, DCR-ReID~\cite{DCR-ReID_tcsvt}, IMS+GEP~\cite{graph}, AFL~\cite{AFL}, CVSL~\cite{cvsl}, MCSC~\cite{meta}, MAL-F~\cite{MAL-F}, and DLCR~\cite{dclr}). 

\begin{table*}[t]
	\centering
	\caption{\textbf{Comparison of R-1 accuracy (\%) and mAP (\%) of state-of-the-art methods on \textsc{LTCC}, \textsc{PRCC}, and \textsc{VC-Clothes}.} Bold and underline values indicate the best and second-best results, respectively. $\dag$ indicates reproduced results.}
	\setlength{\tabcolsep}{7pt}
	\begin{tabular}{lc|cccc|cccc|cccc}
	\toprule[1.1pt]
	\multirow{3}[4]{*}{Methods} & \multirow{3}[4]{*}{Venue} & \multicolumn{4}{c|}{\textsc{LTCC}} & \multicolumn{4}{c|}{\textsc{PRCC}} & \multicolumn{4}{c}{\textsc{VC-Clothes}} \\
	\cmidrule(lr){3-6} \cmidrule(lr){7-10} \cmidrule(lr){11-14}
	& & \multicolumn{2}{c}{General} & \multicolumn{2}{c|}{Cloth-changing} & \multicolumn{2}{c}{Same-clothes} & \multicolumn{2}{c|}{Cloth-changing} & \multicolumn{2}{c}{General} & \multicolumn{2}{c}{Cloth-changing} \\
	\cmidrule(lr){3-4} \cmidrule(lr){5-6} \cmidrule(lr){7-8} \cmidrule(lr){9-10} \cmidrule(lr){11-12} \cmidrule(lr){13-14}
	& & R-1 & mAP & R-1 & mAP & R-1 & mAP & R-1 & mAP & R-1 & mAP & R-1 & mAP \\
	\midrule
	ISP~\cite{isp} & ECCV'20 & 66.3 & 29.6 & 27.8 & 11.9 & 92.8 & - & 36.6 & - & 94.5 & 94.7 & 72.0 & 72.1 \\
	FSAM~\cite{fsam} & CVPR'21 & 73.2 & 35.4 & 38.5 & 16.2 & - & - & - & - & 94.7 & \underline{94.8} & 78.6 & 78.9 \\
	RCSANet~\cite{rcsanet} & ICCV'21 & - & - & - & - & \textbf{100.0} & 97.2 & 50.2 & 48.6 & - & - & - & - \\
	GI-ReID~\cite{gireid} & CVPR'22 & 63.2 & 29.4 & 23.7 & 10.4 & - & - & - & - & - & - & 64.5 & 57.8 \\
	Pos-Neg~\cite{posneg} & TIP'22 & 75.7 & 37.0 & 36.2 & 14.4 & - & - & 54.9 & 65.8 & - & - & - & - \\
	CAL~\cite{cal} $\dag$ & CVPR'22 & 73.4 & 39.4 & 38.0 & 17.2 & \textbf{100.0} & 99.6 & 54.7 & 55.4 & 93.1 & 88.3 & 82.6 & 81.7 \\
	M2Net~\cite{m2net} & ACM MM'22 & - & - & - & - & 99.5 & 99.1 & 59.3 & 57.7 & - & - & - & - \\
	ACID~\cite{tip_win} & TIP'23 & 65.1 & 30.6 & 29.1 & 14.5 & 99.1 & 99.0 & 55.4 & \textbf{66.1} & \underline{95.1} & 94.7 & 84.3 & 74.2 \\
	CLIP-ReID~\cite{clipreid} $\dag$ & AAAI'23 & 73.0 & 36.5 & 38.3 & 15.9 & \textbf{100.0} & 99.5 & 57.0 & 55.3 & 93.1 & 86.9 & 85.9 & 79.3 \\
	3DInvarReID~\cite{3DInvarReID} & ICCV'23 & - & -& 40.9 & 18.9 & - & - & 56.5 & 57.2 & - & - & - & - \\
	DCR-ReID~\cite{DCR-ReID_tcsvt} & TCSVT'23 & 76.1 & 42.3 & 41.1 & 20.4 & \textbf{100.0} & 99.7 & 57.2 & 57.4 & - & - & - & - \\
	AIM~\cite{aim} $\dag$ & CVPR'23 & 75.7 & 41.3 & 41.8 & 17.9 & \textbf{100.0} & 99.7 & 56.3 & 56.5 & 91.2 & 84.9 & 81.0 & 75.7 \\
	Instruct-ReID~\cite{Instruct-ReID} & CVPR'24 & 75.8 & \textbf{52.0} & - & - & - & - & 54.2 & 52.3 & - & - & \underline{89.7} & 78.9 \\
	AFL~\cite{AFL} & TMM'24 & 74.4 & 39.1 & 42.1 & 18.4 & \textbf{100.0} & 99.7 & 57.4 & 56.5 & 93.9 & 88.3 & 82.5 & 83.0 \\
	MCSC~\cite{meta} & TIP'24 & 73.9 & 40.2 & 42.2 & 19.4 & 99.8 & \textbf{99.8} & 57.8 & 57.3 & 93.2 & 87.9 & 83.3 & \underline{83.2} \\
	IMS+GEP~\cite{graph} & TMM'24 & - & - & 43.4 & 18.2 & 99.7 & \textbf{99.8} & 57.3 & \underline{65.8} & 94.7 & \textbf{94.9} & 81.8 & 81.7 \\
	CVSL~\cite{cvsl} & WACV'24 & \underline{76.4} & 41.9 & \underline{44.5} & \underline{21.3} & 97.5 & 99.1 & 57.5 & 56.9 & - & - & - & - \\
	MAL-F~\cite{MAL-F} & TITS'25 & 37.5 & 16.2 & 38.8 & 17.0 & - & - & 57.1 & 57.7 & - & - & - & - \\ 
	DLCR~\cite{dclr} & WACV'25 & - & - & 41.3 & 19.6 & - & - & \textbf{66.5} & {63.0} & - & - & 87.1 & 81.1 \\ 
	SCI \textit{w/} ResNet-50 & Ours & 75.7 & 40.6 & 42.1 & 18.6 & 99.6 & 97.7 & 59.8 & 56.2 & {94.9} & 89.2 & 89.2 & 83.1 \\
	SCI \textit{w/} ViT-B & Ours & \textbf{79.9} & \underline{44.4} & \textbf{45.9} & \textbf{21.6} & \textbf{100.0} & 99.4 & \underline{62.8} & 60.0 & \textbf{95.4} & 91.6 & \textbf{91.7} & \textbf{87.5} \\
	\bottomrule[1.1pt]
	\end{tabular}
	\label{tab_sota}
\end{table*}

As shown in \cref{tab_sota}, SCI consistently outperforms baselines by effectively suppressing ID-irrelevant features and reinforcing discriminative ID cues. 

\subsubsection{CNN-Based Model}

On \textsc{LTCC} (cloth-changing), SCI improves Rank-1 by 3.8\% and mAP by 2.7\% over CLIP-ReID~\cite{clipreid}. Although SCI’s mAP is marginally lower than AIM~\cite{aim} in the general setting, this is expected: by filtering out clothing information, SCI may sacrifice some performance when clothing cues are reliable, but gains robustness under cloth variation. 

On \textsc{PRCC}, SCI surpasses AIM by 3.5\% in Rank-1 accuracy under clothing changes. Due to \textsc{PRCC}’s simpler two-outfit per ID setup, improvements on other metrics are relatively smaller but still notable. 

On \textsc{VC-Clothes}, SCI achieves a substantial mAP gain of 7.4\% over AIM in the cloth-changing setting, further validating its effectiveness in challenging scenarios. 

\subsubsection{ViT-Based Model} 

With ViT-B/16, SCI attains near state-of-the-art performance, especially on \textsc{LTCC}, reaching 45.9\% Rank-1 under severe clothing variations. Crucially, these results are obtained without auxiliary models or additional preprocessing, distinguishing SCI from methods that rely on external components. 

\subsection{Ablation Studies and Analysis}

We perform ablation studies to assess each module’s contribution and analyze key parameters.

\begin{table*}[t]
	\centering
	\caption{\textbf{Ablation study of each component of the SCI on \textsc{LTCC}, \textsc{PRCC}, and \textsc{VC-Clothes}.} Bold values indicate the best results.}
	\setlength{\tabcolsep}{6.5pt}
	\begin{tabular}{cccc|cccc|cccc|cccc} 
	\toprule[1.1pt]
	\multirow{3}[4]{*}{Backbone} & \multirow{3}[4]{*}{Baseline} & \multirow{3}[4]{*}{SSE} & \multirow{3}[4]{*}{SIM} & \multicolumn{4}{c|}{\textsc{LTCC}} & \multicolumn{4}{c|}{\textsc{PRCC}} & \multicolumn{4}{c}{\textsc{VC-Clothes}} \\
	\cmidrule(lr){5-8} \cmidrule(lr){9-12} \cmidrule(lr){13-16}
	& & & & \multicolumn{2}{c}{General} & \multicolumn{2}{c|}{Cloth-changing} & \multicolumn{2}{c}{Same-clothes} & \multicolumn{2}{c|}{Cloth-changing} & \multicolumn{2}{c}{General} & \multicolumn{2}{c}{Cloth-changing} \\
	\cmidrule(lr){5-6} \cmidrule(lr){7-8} \cmidrule(lr){9-10} \cmidrule(lr){11-12} \cmidrule(lr){13-14} \cmidrule(lr){15-16}
	& & & & R-1 & mAP & R-1 & mAP & R-1 & mAP & R-1 & mAP & R-1 & mAP & R-1 & mAP \\
	\midrule
	\multirow{4}{*}{ResNet-50} & \CIRCLE & \Circle & \Circle & 73.0 & 36.5 & 38.3 & 15.9 & \textbf{100.0} & \textbf{99.5} & 57.0 & 55.3 & 93.1 & 86.9 & 85.9 & 79.3 \\
	& \CIRCLE & \CIRCLE & \Circle & 76.5 & 39.7 & 39.8 & 18.3 & \textbf{100.0} & 98.4 & 59.0 & 54.7 & 94.1 & 88.0 & 87.6 & 81.1 \\
	& \CIRCLE & \Circle & \CIRCLE & 72.8 & 36.3 & 37.8 & 16.1 & 99.7 & 98.0 & 52.4 & 49.9 & 94.0 & 88.9 & 88.4 & 82.4 \\
	
	& \CIRCLE & \CIRCLE & \CIRCLE & \textbf{75.7} & \textbf{40.6} & \textbf{42.1} & \textbf{18.6} & 99.6 & 97.7 & \textbf{59.8} & \textbf{56.2} & \textbf{94.9} & \textbf{89.2} & \textbf{89.2} & \textbf{83.1} \\
	
	\midrule
	\multirow{4}{*}{ViT-B} & \CIRCLE & \Circle & \Circle & 79.5 & 43.6 & 41.6 & 19.4 & 99.9 & 99.2 & 60.9 & 58.5 & 94.4 & 89.8 & 90.6 & 84.7 \\
	& \CIRCLE & \CIRCLE & \Circle & 76.3 & 43.9 & 43.4 & 21.0 & 100.0 & 99.3 & 62.3 & 58.7 & 94.6 & 89.8 & 91.4 & 84.8 \\
	& \CIRCLE & \Circle & \CIRCLE & 77.5 & 43.9 & 42.9 & 20.5 & 99.8 & 99.2 & 61.4 & 59.1 & 94.3 & 90.4 & 90.9 & 85.7 \\
	& \CIRCLE & \CIRCLE & \CIRCLE & \textbf{79.9} & \textbf{44.4} & \textbf{45.9} & \textbf{21.6} & \textbf{100.0} & \textbf{99.4} & \textbf{62.8} & \textbf{60.0} & \textbf{95.4} & \textbf{91.6} & \textbf{91.7} & \textbf{87.5} \\
	\bottomrule[1.1pt]
	\end{tabular}
	\label{tab_ltcc}
\end{table*}

\subsubsection{Impact of Semantic Separation Enhancement}

We first evaluate the baseline without additional modules, where the model relies solely on a prompt ``\texttt{a photo of a $[X]_1 \dots [X]_M$ person}'' for classification. This approach shows reduced performance. Introducing the SSE module, which uses two prompts (``\texttt{a photo of a $[X]_1 \dots [X]_M$ person}'' and ``\texttt{a photo of the $[X]_1 \dots [X]_M$ clothes}'') to learn features separately at the textual level. We observe consistent improvements of approximately 2.0\% in both Rank-1 accuracy and mAP across datasets for both CNN-based and ViT-based models, as shown in \cref{tab_ltcc}. This demonstrates the effectiveness of the SSE module in filtering clothing features.


These results indicate: 
\textit{1)} The original generic textual representations are confounding, containing both ID and clothing features, leading the model to focus excessively on cloth-relevant aspects.
\textit{2)} Separating these features at the textual level is simple yet effective.
These findings demonstrate the effectiveness of setting specific prompts for each factor and the robustness of our SSE module against both cloth-changing and general scenarios.

\subsubsection{Impact of Semantic-Guided Interaction Module}

We further assess SIM by integrating it with the baseline model. As shown in \cref{tab_ltcc}, using SIM alone often degrades performance based on CNN, particularly in the cloth-changing scenario on \textsc{PRCC}, where Rank-1 drops by 4.6\% and mAP decreases from 55.3\% to 49.9\%.

The decline in performance in this set of ablation experiments is primarily due to the absence of the SSE module. Typically, the SSE module generates final textual features that guide the visual representations during the updating process of the visual encoder in SIM. However, with only a single prompt ``\texttt{a photo of a $[X]_1 \dots [X]_M$ person}'' used to extract textual representations, many ID-irrelevant negative factors, particularly clothing factors, are included. This results in an excessive emphasis on clothing-relevant representations during guidance in SIM, reducing overall performance.


\begin{figure*}[t]
 	\centering
 	\begin{tabular}{cccc}
 	\includegraphics[width = 0.22\linewidth]{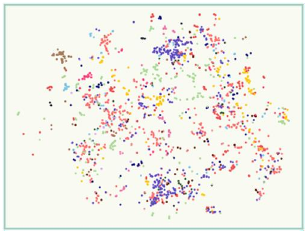} & 
 	\includegraphics[width = 0.22\linewidth]{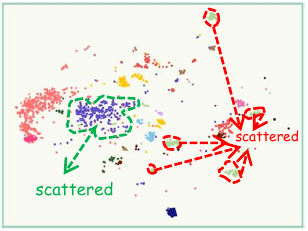} & 
 	\includegraphics[width = 0.22\linewidth]{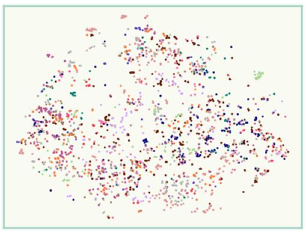} & 
 	\includegraphics[width = 0.22\linewidth]{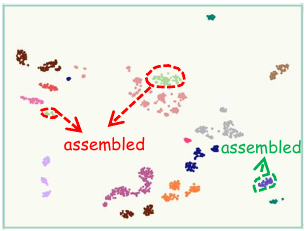} \\
 	\footnotesize{(a) Baseline (stage1)} & \footnotesize{(b) Baseline (stage2)} & \footnotesize{(c) Ours (stage1)} & \footnotesize{(d) Ours (stage2)}
 	\end{tabular}
 	\caption{\textbf{t-SNE visualization of 20 randomly selected classes from \textsc{LTCC}.} Colors indicate ground-truth IDs. (a)--(b) depict successive stages of the baseline, while (c)--(d) show the corresponding stages of our method.}
 	\label{fig_tsne}
\end{figure*}

\subsubsection{Feature Distribution via t-SNE}

To illustrate SCI's effectiveness, we employ t-SNE visualization~\cite{tSNE}, as depicted in \cref{fig_tsne}. This visualization plots the statistical distribution for 20 randomly selected categories from \textsc{LTCC}, comparing the latent space distributions at different stages of the baseline model with our approach.

From \cref{fig_tsne}(a) and (c), it is observable that both the baseline and our method exhibit disorganized feature distributions with blurred class boundaries. During this stage, only the prompts are trained while the visual and text encoders remain frozen, limiting their capability to effectively extract diverse representations. In contrast, as shown in \cref{fig_tsne}(b) and (d), the feature distributions for both methods become clearer, indicating the effectiveness of the representations derived from the text-image similarity computations.

Notably, the clusters in \cref{fig_tsne}(d) are more compact and distinct compared to those in \cref{fig_tsne}(b). The red and green dashed circles in \cref{fig_tsne}(b) enclose samples with considerable scatter, which are better clustered together in \cref{fig_tsne}(d). Therefore, this visualization not only serves as compelling validation of our capability to effectively extract features and discriminate IDs but also highlights our method's potential to significantly advance the field in cloth-changing scenarios.

\begin{figure}[t]
	\centering
	\includegraphics[width = \linewidth]{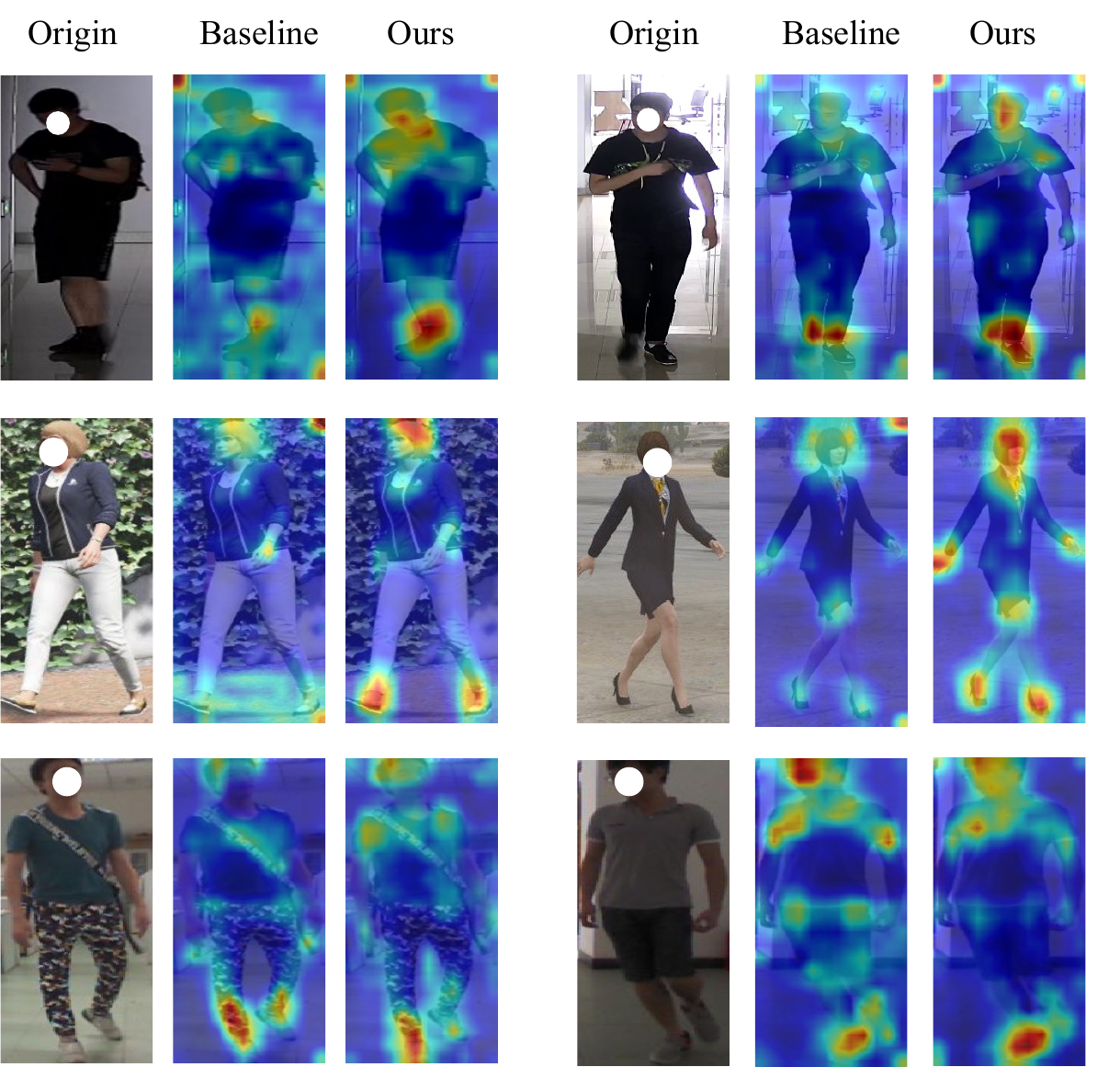}
	\caption{\textbf{Visualization of feature maps on \textsc{LTCC} (first row), \textsc{VC-Clothes} (second row), and \textsc{PRCC} (third row).} The first column shows the original images, while the second and third columns present the feature maps from the baseline and our method, respectively.}
	\label{fig_gradcam}
\end{figure}

\subsubsection{Qualitative Prompt-Guided Refinement}

To illustrate the distinctions between the baseline~\cite{clipreid} and SCI, we present in \cref{fig_gradcam} the attention maps activated by different prompts, highlighting how prompts guide the refinement of visual features across \textsc{LTCC}, \textsc{VC-Clothes}, and \textsc{PRCC}. It demonstrates that the learned prompts are capable of guiding the network to focus on relevant semantic areas. 

\begin{figure}[t]
	\centering
	\includegraphics[width = \linewidth]{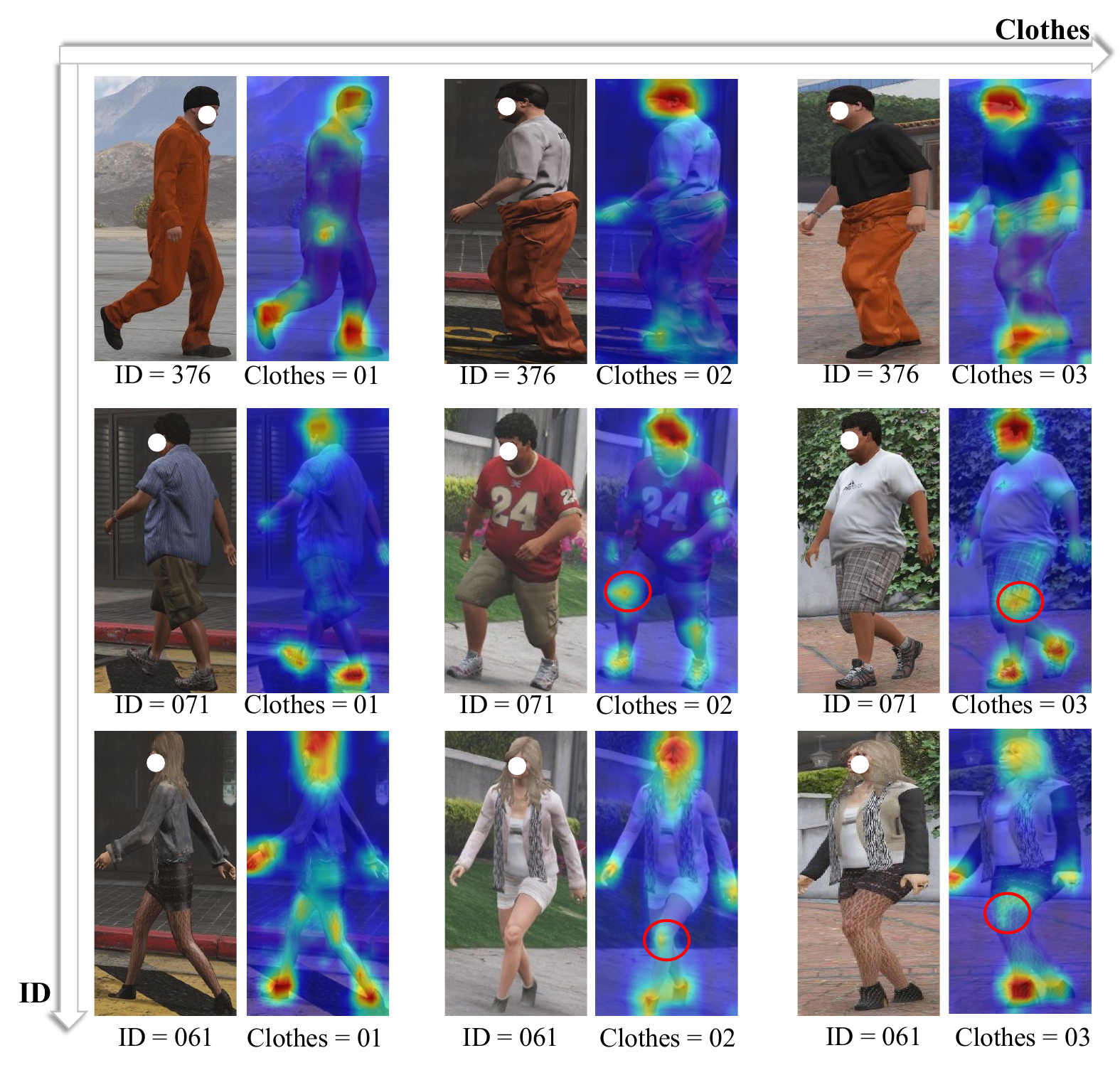}
	\caption{\textbf{Visualization of feature maps on \textsc{VC-Clothes}.} The vertical axis represents different IDs, and the horizontal axis shows clothing variations. Red circles highlight challenging cases. This figure illustrates the model’s attention to the same ID across different clothing scenarios.}
	\label{fig_sameID}
\end{figure}

\cref{fig_sameID} shows \textsc{VC-Clothes} for the same ID but with different outfits, illustrating the regions of interest that our SCI model focuses on under clothing variations. These visualizations provide insights into the unique differences in feature activation patterns. Moreover, while SCI leverages prompt learning to jointly model ID and clothing information, it fundamentally differs from attribute-based approaches, as it neither explicitly recognizes attributes nor relies on additional attribute detectors.

As shown in \cref{fig_gradcam}, the baseline method tends to focus on more dispersed areas, which include negative factors, potentially detracting from its effectiveness. Conversely, SCI emphasizes highly discriminative ID features, such as the person's head, shoulders, lower body, and shoes. This specific emphasis on footwear aligns with findings from previous studies~\cite{maskattribute}, which suggest that footwear often remains consistent across different outfits and scenarios. This indicates our method’s enhanced ability to capture crucial positive factors when encountering challenging scenarios.

In \cref{fig_sameID}, for each person, the corresponding heatmaps illustrate how the feature extraction model highlights various regions of interest across different outfits, reflecting the model’s focus on certain body parts regardless of clothing changes. However, there are some limitations. As indicated by the red circles, the model tends to focus excessively on areas like the knees. If people wear long pants, this focus could potentially affect the results. We are aware of this issue and will continue to optimize our model to address it.

\begin{figure*}[t]
 	\centering
 	\begin{tabular}{ccc}
 	\includegraphics[width = 0.3\linewidth]{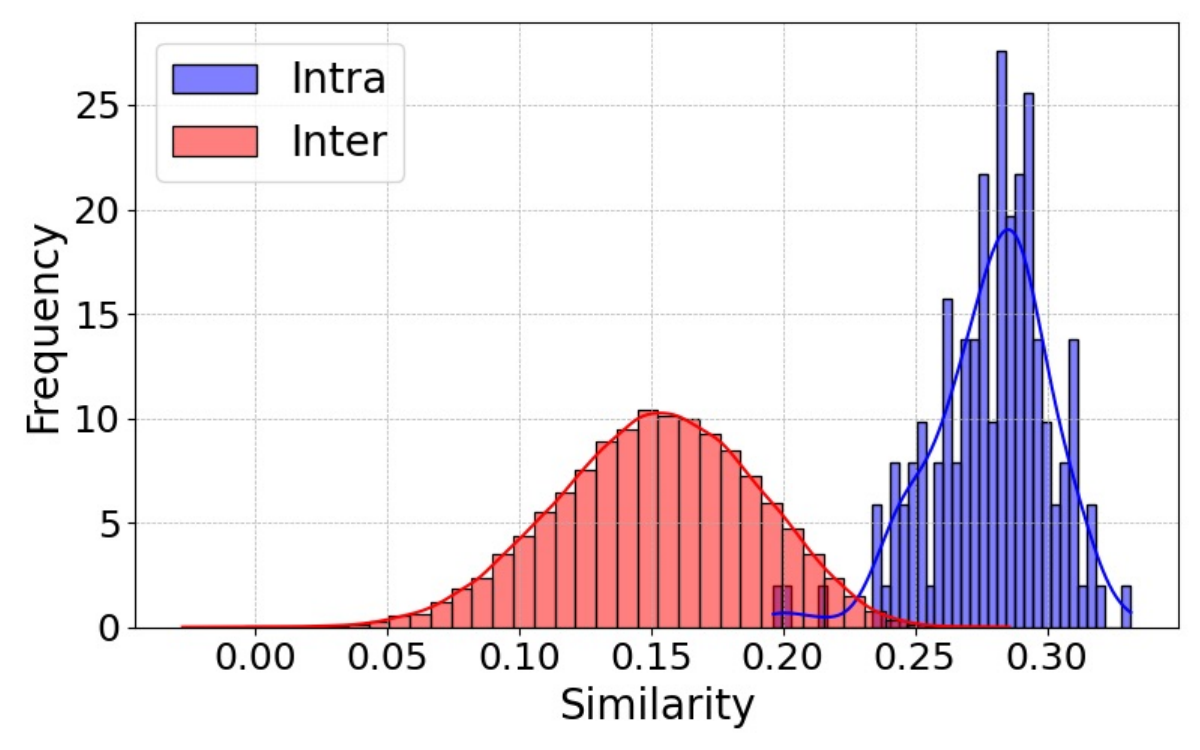} & \includegraphics[width = 0.3\linewidth]{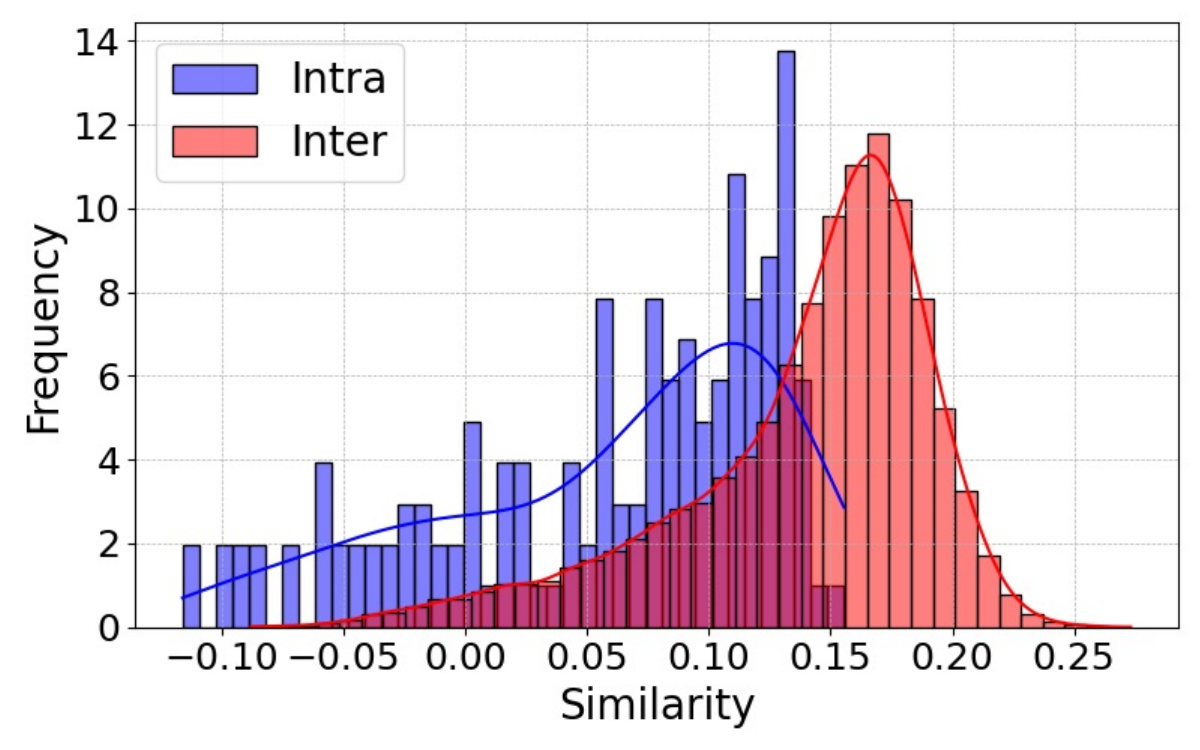} & 
 \includegraphics[width = 0.3\linewidth]{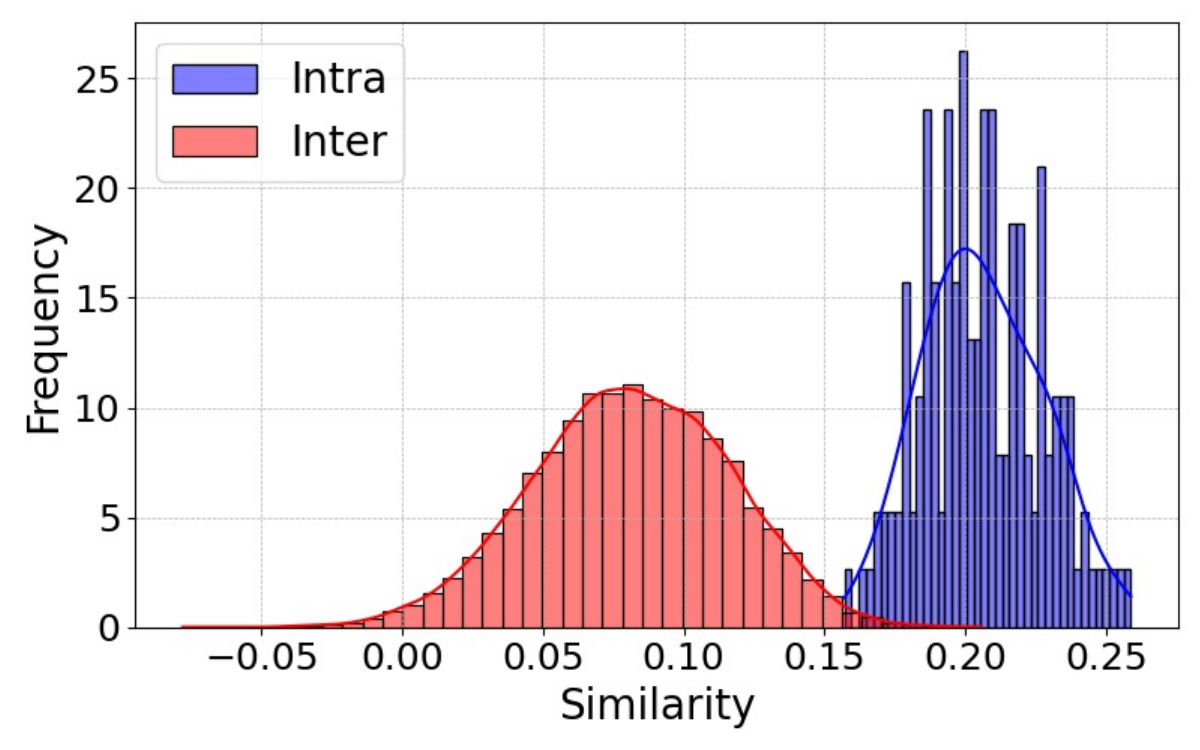}\\
 	\footnotesize{(a) $F^\mathrm{txt}_\mathrm{id}$} distribution. & \footnotesize{(b) $F^\mathrm{txt}_\mathrm{clo}$} distribution. & \footnotesize{(c) $F^\mathrm{txt}_\mathrm{ort}$ distribution.} 
 	\end{tabular}
 	\caption{\textbf{Visualization of similarity distributions on \textsc{PRCC}.} (a)--(c) show the similarity distributions between image features and three different types of text features.}
 	\label{fig_similarity}
\end{figure*}

\subsubsection{Visualization of Feature Similarity}
In \cref{fig_similarity}, we visualize the similarity distributions of features between the text prompts and the images. \cref{fig_similarity}(a) is derived from the prompt in \cref{eq_learncontext1}, while the distribution is relatively well-formed, intra-class similarity remains dispersed. \cref{fig_similarity}(b) shows the distribution from the prompt in \cref{eq_learncontext2}, which learns clothing-related semantics. However, due to significant clothing variations over time for the same individual, intra-class differences outweigh inter-class differences. \cref{fig_similarity}(c) shows the distribution after orthogonalizing the features, which achieves a clearer separation between intra- and inter-class similarity, indicating improved feature discrimination. This demonstrates the effectiveness of dual textual prompts and orthogonalization in enhancing feature alignment.

\begin{figure*}[t]
 	\centering
 	\begin{tabular}{cccc}
 	\includegraphics[width = 0.22\linewidth]{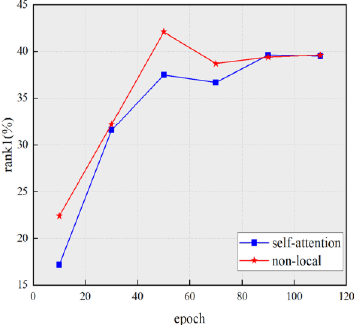} & 
 	\includegraphics[width = 0.22\linewidth]{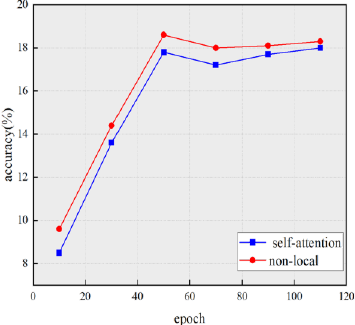} & 
 	\includegraphics[width = 0.22\linewidth]{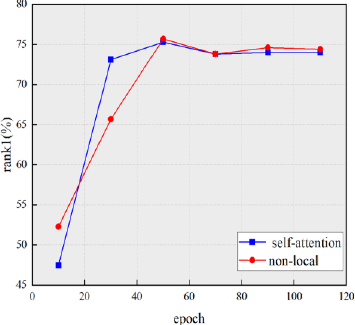} & 
 	\includegraphics[width = 0.22\linewidth]{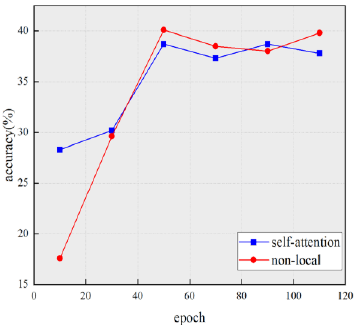} \\
 	\footnotesize{(a) R-1 on cloth-changing} & \footnotesize{(b) mAP on cloth-changing} & \footnotesize{(c) R-1 on general} & \footnotesize{(d) mAP on general}
 	\end{tabular}
 	\caption{\textbf{Evaluation of SIM module operations on \textsc{LTCC}.} (a) and (b) show the CMC curves and mAP (\%) for cloth-changing scenarios, while (c) and (d) present the CMC curves and mAP (\%) for general scenarios, demonstrating the superior performance of our method.}
 	\label{fig_nonlocal}
\end{figure*}

\subsubsection{Comparative Analysis of Mechanisms in SIM}

As shown in \cref{fig_nonlocal}, we evaluate the effectiveness of the non-local operation~\cite{nonlocal} compared to self-attention mechanisms~\cite{attentionisall} in leveraging textual information obtained from the SSE module to guide visual representations. In cloth-changing scenarios, the non-local mechanism yields superior performance, achieving 42.1\% in Rank-1 accuracy and 18.6\% in mAP, as depicted in \cref{fig_nonlocal}(a) and (b), thereby outperforming the self-attention mechanism by margins of 4.6\% and 2.4\%, respectively.

This underscores the non-local mechanism’s enhanced capability to capture comprehensive global context, which is vital for integrating information from disparate yet significant segments of the visual field in cloth-changing scenarios. In general scenarios, as illustrated in \cref{fig_nonlocal}(c) and (d), both mechanisms exhibit comparable efficacy, with the non-local mechanism slightly edging out self-attention by a 2.0\% increase in mAP. This suggests the non-local mechanism is robust and can be effectively deployed for person ReID tasks.

\begin{figure}[t]
	\centering
	\begin{tabular}{cc}
	\includegraphics[width = 0.45\linewidth]{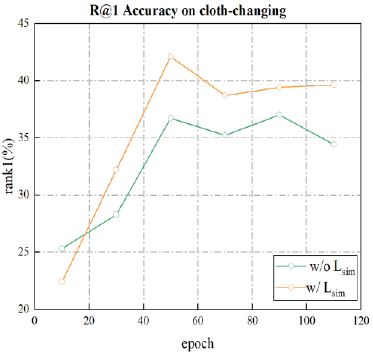} & 
	\includegraphics[width = 0.45\linewidth]{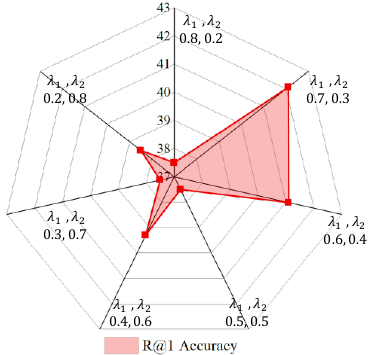} \\
	\footnotesize{(a) ablation analysis of \cref{eq_lambda}} & \footnotesize{(b) $\lambda_1$, $\lambda_2$ analysis}
	\end{tabular}
	\caption{\textbf{Ablation studies on loss function and parameter effects.} (a) presents the ablation analysis of \cref{eq_lambda} with Rank-1 (\%). (b) shows the parameter analysis on \textsc{LTCC} for the weighting factors $\lambda_1$ and $\lambda_2$ in \cref{eq_lambda}.}
	\label{fig_loss}
\end{figure}

\subsubsection{Parameter Sensitivity}

We conduct experimental evaluations to explore the effectiveness of \cref{eq_lambda}, as demonstrated in \cref{fig_loss}(a). This loss function significantly improves performance by optimizing for higher ID similarity and reducing clothing similarity, making it particularly effective for applications that need to distinguish personal ID from external features like clothing, relevant in tasks such as cloth-changing scenarios. It helps the model focus on positive factors while ignoring negative variations such as clothing changes, thereby enhancing adaptability and accuracy in complex environments.

\subsubsection{Generalization to Cloth-Consistent Scenarios}

To validate the applicability of our method in cloth-consistent scenarios, we conducted experiments using our cloth-changing approach on two traditional ReID datasets: \textsc{Market1501} and \textsc{MSMT17}. For ease of comparison, we also incorporated cloth-consistent methods: AGW~\cite{agw}, TransReID~\cite{transreid}, CLIP-ReID~\cite{clipreid}, and Instruct-ReID~\cite{Instruct-ReID}. This enables a comprehensive evaluation of the performance in varying scenarios.

\begin{table}[t]
	\centering
	\caption{\textbf{Comparison of R-1 accuracy (\%) and mAP (\%) of state-of-the-art methods on \textsc{Market1501} and \textsc{MSMT17} on two categories of methods.} $\dag$ indicates reproduced results. Bold values indicate the best results.}
	\setlength{\tabcolsep}{3pt}
	\begin{tabular}{clc|cc|cc}
	\toprule[1.1pt]
	& \multirow{2}[2]{*}{Methods} & \multirow{2}[2]{*}{Venue} & \multicolumn{2}{c|}{\textsc{Market1501}} & \multicolumn{2}{c}{\textsc{MSMT17}} \\
	\cmidrule(lr){4-5} \cmidrule(lr){6-7} 
	& & & R-1 & mAP & R-1 & mAP \\
	\midrule

	\multirow{5}{*}{\makecell{Cloth-\\consistent}}
	& AGW~\cite{agw} & TPAMI'21 & 95.1 & 87.8 & 68.3 & 49.3 \\
	& TransReID~\cite{transreid} & ICCV'21 & 95.2 & 89.5 & 86.2 & 69.4 \\
	& CLIP-ReID~\cite{clipreid} & AAAI'23 & 95.7 & 89.8 & 84.4 & 63.0 \\
 & CMT~\cite{cmt_tcsvt} & TCSVT'24 & 95.8 & 88.9 & 78.5 & 51.3 \\
	& Instruct-ReID~\cite{Instruct-ReID} & CVPR'24 & \textbf{96.5} & \textbf{93.5} & \textbf{86.9} & \textbf{72.4} \\

	\midrule
	\multirow{7}{*}{\makecell{Cloth-\\changing}}
	& CAL~\cite{cal} $\dag$ & CVPR'22 & 85.0 & 67.8 & 65.1 & 38.4 \\
	& AIM~\cite{aim} $\dag$ & CVPR'23& 70.3 & 48.0 & 26.6 & 12.0 \\ 
	& 3DInvarReID~\cite{3DInvarReID} & ICCV'23
	& 95.1 & 97.9 & 80.8 & 59.1 \\
	& AFL~\cite{AFL} & TMM'24 & \textbf{95.5} & 88.8 & 80.9 & 58.1 \\
	& FIRe$^2$~\cite{Fire2} & TIFS'24 & 95.4 & 87.7 & 79.7 & 56.2 \\
	& \cellcolor{gray!20}SCI \textit{w/} ResNet-50 & \cellcolor{gray!20} Ours & \cellcolor{gray!20} 94.4 & \cellcolor{gray!20} 86.5 & \cellcolor{gray!20} 81.4 & \cellcolor{gray!20} 59.5 \\
	& \cellcolor{gray!20}SCI \textit{w/} ViT-B & \cellcolor{gray!20} Ours & \cellcolor{gray!20} 95.2 & \cellcolor{gray!20} \textbf{89.4} & \cellcolor{gray!20} \textbf{89.6} & \cellcolor{gray!20} \textbf{74.7} \\
	\bottomrule[1.1pt]
	\end{tabular}
	\label{tab_consistent}
\end{table}

\begin{table*}[t]
	\centering
	\caption{\textbf{Comparison of R-1 accuracy (\%) and efficiency of different methods.} $\dag$ denotes reproduced results.} 
 	\setlength{\tabcolsep}{6pt}
	\begin{tabular}{lcc|cccc|ccc}
	\toprule[1.1pt]
	\multirow{2}[2]{*}{Methods} & \multirow{2}[2]{*}{Venue} & 
	\multirow{2}[2]{*}{Input Resolution} &
	\multirow{2}[2]{*}{Params (M)} & \multirow{2}[2]{*}{GFLOPs} & \multirow{2}[2]{*}{Training Time (min)} & \multirow{2}[2]{*}{FPS} & \multicolumn{3}{c}{R-1} \\
	\cmidrule(lr){8-10}
	& & & & & & & \textsc{LTCC} & \textsc{PRCC} & \textsc{VC-Clothes} \\
	\midrule
	CAL~\cite{cal} $\dag$ & CVPR'22 & $384 \times 192$ & 23.5 & 9.2 & 41 & 538 & 38.0 & 54.7 & 82.6 \\
	CLIP-ReID~\cite{clipreid} $\dag$ & AAAI'23 & $384 \times 192$ & 77.6 & 10.6 & 67 & 314 & 38.3 & 57.0 & 85.9 \\
	3DInvarReID~\cite{3DInvarReID} & ICCV'23 & $384 \times 128$ & 217.6 & -- & 241 & -- & 40.9 & 56.5 & - \\
	AIM~\cite{aim} $\dag$ & CVPR'23 & $384 \times 192$ & 72.7 & 9.2 & 52 & 754 & 41.8 & 56.3 & 81.0 \\
	\rowcolor{gray!20} 
	SCI \textit{w/} ResNet-50 & Ours & $384 \times 192$ & 93.9 & 10.6 & 78 & 302 & 42.1 & 59.8 & 89.2 \\
	\rowcolor{gray!20} 
	SCI \textit{w/} ViT-B & Ours & $384 \times 192$ & 143.7 & 26.7 & 46 & 80 & \textbf{45.9} & \textbf{62.8} & \textbf{91.7} \\
	\bottomrule[1.1pt]
	\end{tabular}
	\label{tab_gflops}
\end{table*}

As shown in \cref{tab_consistent} (cloth-consistent), existing ReID methods designed for traditional datasets have achieved progressively better performance, even on larger-scale datasets like \textsc{MSMT17}. Although our proposed SCI framework is primarily designed to address the more practical scenario of clothing changes, we also evaluate its effectiveness on conventional datasets to assess its generalization ability. 
From \cref{tab_consistent} (cloth-changing), it can be observed that our SCI method achieves the best performance on both \textsc{Market1501} and \textsc{MSMT17} among cloth-changing methods. In particular, the ViT-based SCI achieves a Rank-1 accuracy of 89.6\% on \textsc{MSMT17}, outperforming Instruct-ReID by 2.7\% in Rank-1 and 2.3\% in mAP. These results underscore the strong generalization capability of our approach. 

\begin{figure}[t]
	\centering
	\includegraphics[width = \linewidth]{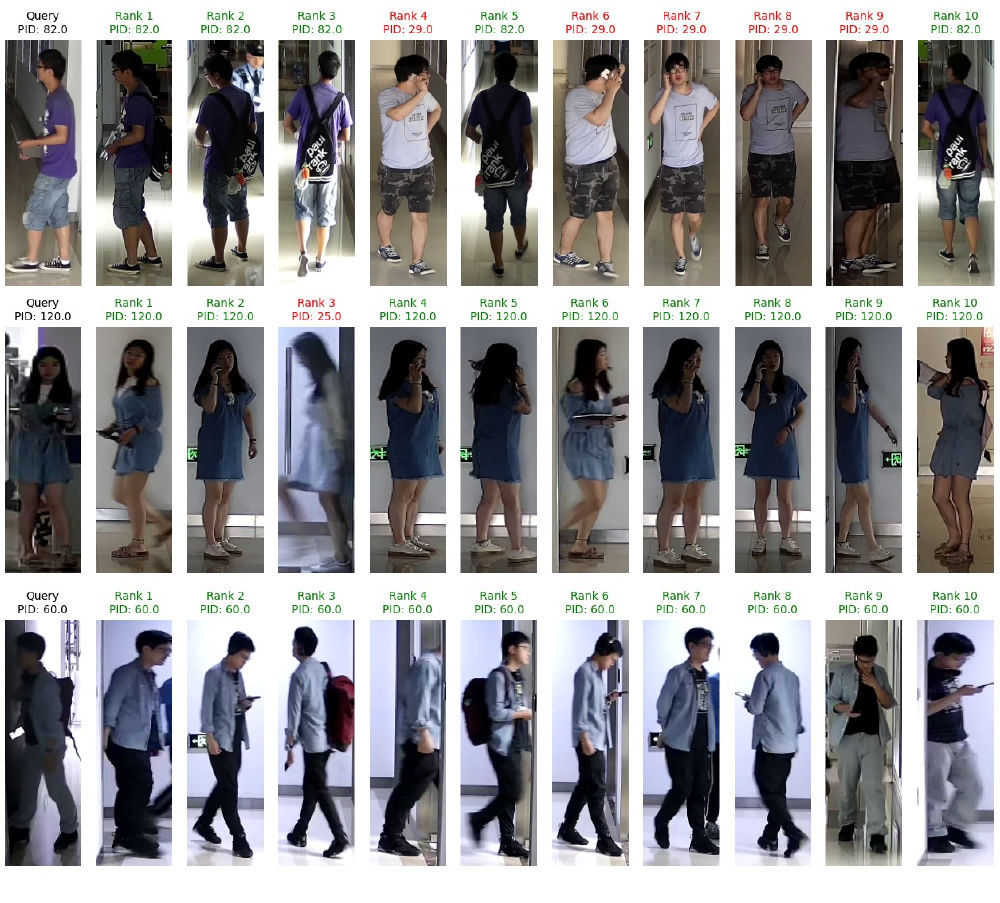}
	\caption{\textbf{Top-10 retrieval results on \textsc{LTCC}.} All retrieved images are from the gallery and captured by different cameras. Correct matches are highlighted in green, while incorrect ones are shown in red. PID denotes the person ID.}
	\label{fig_retrieve_sucess}
\end{figure}

\subsubsection{Visualization of Retrieved Examples}
We present the visualization of the top-10 retrieved results from randomly selected query examples on \textsc{LTCC} in \cref{fig_retrieve_sucess}. It observes that the retrieval results are predominantly optimistic, reflecting a high level of accuracy and relevance in most cases. 

\begin{figure}[t]
	\centering
	\includegraphics[width = \linewidth]{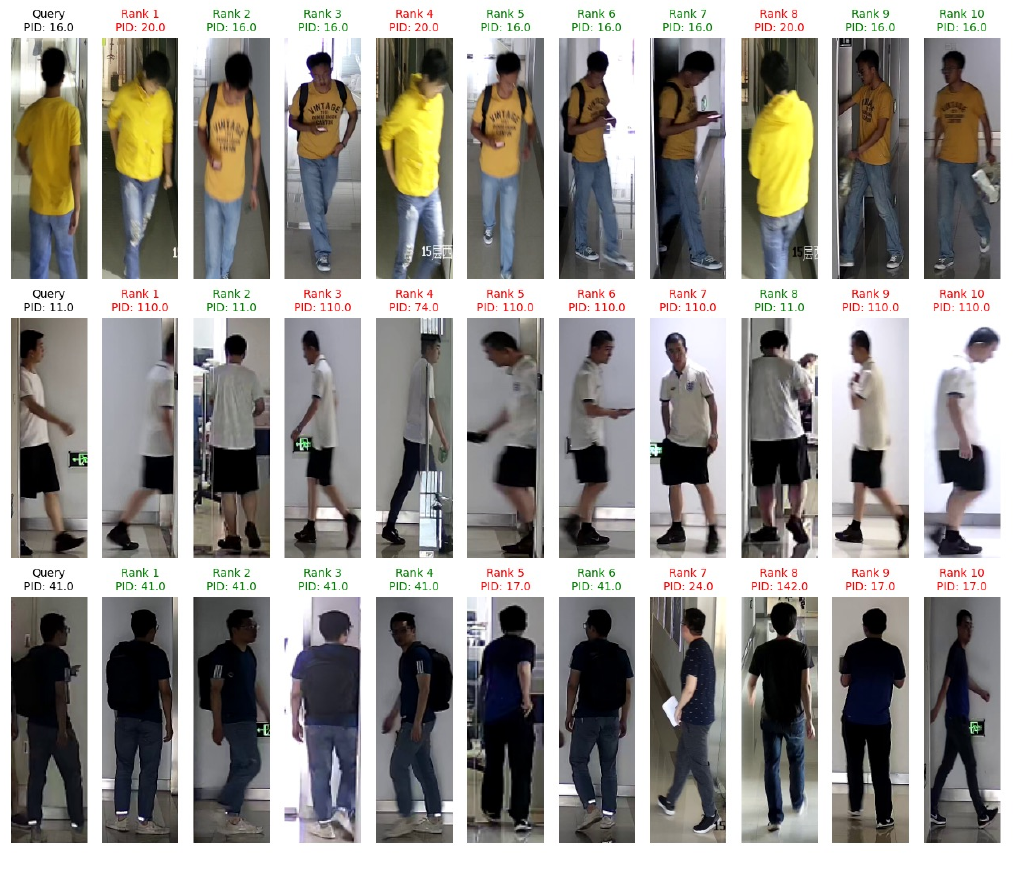}
	\caption{\textbf{Failure cases on \textsc{LTCC}.} All retrieved images are from the gallery and captured by different cameras. Correct matches are highlighted in green, while incorrect ones are shown in red. PID denotes the person ID.}
	\label{fig_retrieve_fail}
\end{figure}

However, some incorrect results still persist, as shown in \cref{fig_retrieve_fail}, mainly due to limited views and highly similar clothing colors, which make retrieval harder. This problem is common when people wear uniforms or dark winter clothes. Further improvements are needed to enhance the model’s ability to distinguish individuals with similar appearances.

\subsubsection{Parameter Selection}
We evaluate the parameters $\lambda_1$ and $\lambda_2$ specified in \cref{eq_lambda}. The results for \textsc{LTCC} are displayed in \cref{fig_loss}(b). We determine that $\lambda_1 = 0.7$ and $\lambda_2 = 0.3$ provide the best performance in most scenarios, optimizing the balance between enhancing ID and reducing clothing similarity, while feature orthogonalization minimizes interference between ID and clothing features. This setting allows the model to effectively discern between the positive representations and their negative factors, such as clothing. Additionally, $\lambda_1 = 0.6$ and $\lambda_2 = 0.4$ obtain the second-best results. This analysis underscores the effectiveness and adaptability of our proposed method in real-world surveillance applications.

\subsubsection{Efficiency Comparison}

\cref{tab_gflops} presents a comprehensive comparison of parameter size, computational complexity (GFLOPs), training time, inference speed (FPS), and Rank-1 accuracy across representative ReID methods. 

AIM~\cite{aim} utilizes a dual-branch ResNet model, achieving an Rank-1 accuracy of 41.8\%, which surpasses CAL~\cite{cal}, another CNN-based model. Although AIM requires higher parameters compared to CAL, it significantly outperforms CAL in terms of accuracy with the same parameter count.

In CLIP-ReID~\cite{clipreid} and our method, we integrate textual modality information, transitioning from a single-modality to a multimodal approach. This justifies the observed increases in both GFLOPs and parameters, which remain within acceptable limits. Our SCI \textit{w/} ResNet-50 incorporates a prompt learning strategy and achieves 42.1\% Rank-1 on \textsc{LTCC}, surpassing CLIP-ReID by 3.8\%, while maintaining similar computational efficiency (10.6 GFLOPs vs. 10.6, and 302 FPS vs. 314). This highlights the effectiveness of our design in utilizing textual guidance without introducing significant overhead. 
Furthermore, our SCI \textit{w/} ViT-B achieves the best performance across all datasets, achieving 45.9\% Rank-1 accuracy on \textsc{LTCC}, 62.8\% on \textsc{PRCC}, and 91.7\% on \textsc{VC-Clothes}. Although it has higher computational requirements due to the ViT backbone (143.7M parameters, 26.7 GFLOPs), the performance gains are substantial. 
Overall, SCI achieves a good trade-off between accuracy and efficiency, making it well-suited for practical deployment in real-world scenarios.

\section{Conclusion and Limitation}

In this study, we propose the Semantic Contextual Integration (SCI) network, a novel method that incorporates prompt learning to address cloth-changing person re-identification (CC-ReID). The SCI network includes a Semantic Separation Enhancement (SSE) module that generates ID-specific and clothing prompts to effectively isolate clothing semantics (negative factors) from ID information, thereby preserving high-dimensional ID cues (positive factors) in visual features. Additionally, the Semantic-Guided Interaction Module (SIM) leverages these textual cues to guide and enhance the discrimination of visual features, facilitating seamless interaction between the visual and textual branches. Our work provides valuable insights and suggests promising directions for future research on the role of prompt learning in CC-ReID.

Although learnable prompts outperform the manually crafted prompts used in CLIP, they often lack semantic clarity and remain uninterpretable to human observers. Consequently, the learned prompts cannot be directly visualized in a human-readable format. Future research should address the interpretability of these prompts and develop techniques to enhance their comprehensibility without compromising their discriminative power.

\bibliographystyle{IEEEtran}
\bibliography{CCREID}

\end{document}